\documentclass[conference]{IEEEtran}
\IEEEoverridecommandlockouts

\usepackage{cite}
\usepackage{amsmath,amssymb,amsfonts}
\usepackage{algorithmic}
\usepackage[bookmarks=false]{hyperref}
\usepackage{graphicx}
\usepackage{subcaption}

\graphicspath{ {./figures/} }
\usepackage{tikz}
\usetikzlibrary{shapes.geometric, arrows}
\usepackage{textcomp}
\usepackage{xcolor}
\usepackage{siunitx}
\def\BibTeX{{\rm B\kern-.05em{\sc i\kern-.025em b}\kern-.08em
    T\kern-.1667em\lower.7ex\hbox{E}\kern-.125emX}}
    
\usepackage{url}

\usepackage{array,multirow}

\usepackage{textcomp}
\usepackage{lipsum}
\newcommand\copyrighttext{%
    \footnotesize \textcopyright  2019 IEEE.  Personal use of this material is permitted.  Permission from IEEE must be obtained for all other uses, in any current or future media, including reprinting/republishing this material for advertising or promotional purposes, creating new collective works, for resale or redistribution to servers or lists, or reuse of any copyrighted component of this work in other works.
}
\newcommand\copyrightnotice{%
    \begin{tikzpicture}[remember picture,overlay]
    \node[anchor=south,yshift=10pt, xshift=10pt] at (current page.south) {\fbox{\parbox{\dimexpr\textwidth-\fboxsep-\fboxrule\relax}{\copyrighttext}}};
    \end{tikzpicture}%
}

\begin{document}

\title{A Deep Learning-based Radar and Camera Sensor Fusion Architecture for Object Detection}

\author{
    \IEEEauthorblockN{
        Felix Nobis\IEEEauthorrefmark{1}, Maximilian Geisslinger\IEEEauthorrefmark{2}, Markus Weber\IEEEauthorrefmark{3}, Johannes Betz and Markus Lienkamp}
    \IEEEauthorblockA{
        Chair of Automotive Technology, Technical University of Munich\\
        Munich, Germany \\
        Email: \IEEEauthorrefmark{1}nobis@ftm.mw.tum.de, \IEEEauthorrefmark{2}maximilian.geisslinger@tum.de, \IEEEauthorrefmark{3}markus.weber@tum.de
    }
}

\maketitle
\copyrightnotice


\begin{abstract}
Object detection in camera images, using deep learning has been proven successfully in recent years. Rising detection rates and computationally efficient network structures are pushing this technique towards application in production vehicles. Nevertheless, the sensor quality of the camera is limited in severe weather conditions and through increased sensor noise in sparsely lit areas and at night. Our approach enhances current 2D object detection networks by fusing camera data and projected sparse radar data in the network layers. The proposed CameraRadarFusionNet (CRF-Net) automatically learns at which level the fusion of the sensor data is most beneficial for the detection result. Additionally, we introduce \textit{BlackIn}, a training strategy inspired by Dropout, which focuses the learning on a specific sensor type. We show that the fusion network is able to outperform a state-of-the-art image-only network for two different datasets. The code for this research will be made available to the public at: \href{https://github.com/TUMFTM/CameraRadarFusionNet}{https://github.com/TUMFTM/CameraRadarFusionNet}
\end{abstract}



\section{Introduction}
\begin{figure}[htbp]
	\centering
\includegraphics[width=88mm]{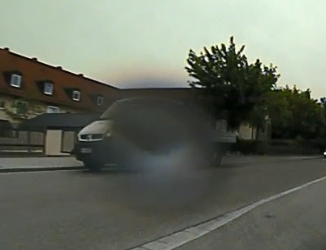}
	\caption{Van occluded by a water droplet on the lens}
	\label{fig:water_droplet_noise}
\end{figure}

In recent years convolutional neural networks (CNN) have been established as the most accurate methods for performing object detection in camera images \cite{Lin.2019}. The visual representation of the environment in camera images is closely linked to human visual perception. As humans perceive the driving environment mainly via their visual sense, it is well motivated for autonomous vehicles to rely on a comparable representation. However, in adverse conditions like heavy rain or fog, the visibility is reduced, and safe driving might not be guaranteed. In addition, camera sensors get increasingly affected by noise in sparsely lit conditions. Compared to camera sensors, radar sensors are more robust to environment conditions such as lighting changes, rain and fog \cite{PonteMuller.2017}. The camera can be rendered unusable through weather-induced occlusion e.g. if water droplets stick to the camera lens and block the view, as shown in Figure~\ref{fig:water_droplet_noise}.

In this paper, we investigate the fusion of radar and camera sensor data with a neural network, in order to increase the object detection accuracy. The radar acquires information about the distance and the radial velocity of objects directly. It is able to locate objects in a two-dimensional plane parallel to the ground. In contrast to the camera, no height information can be obtained by the radar sensor. We develop a network architecture that deals with camera and radar sensor data jointly. The proposed method is able to detect objects more reliably in the nuScenes dataset \cite{Caesar.2019} and the TUM dataset which is created for this research. Additionally, we show the limitations of our fusion network and directions for future development.

Section~\ref{sec:related_work} discusses related methods for object detection and sensor fusion. Section~\ref{sec:radar_processing} describes our method to preprocess the radar data before fusing it into the network. We continue to describe the network architecture in Section~\ref{sec:architecture}. The evaluation and discussion of the approach is performed in Section~\ref{sec:results}. Finally, our conclusions from the work are presented in Section~\ref{sec:conlusions}.


\section{Related Work}
\label{sec:related_work}

\cite{Krizhevsky.2012} were the first to successfully implement a convolutional neural network for the classification of images that outperformed the state-of-the-art in the ImageNet competition. This marked a starting point for increased interest in research into image processing with neural networks. Sebsequently, neural network architectures for classification are augmented to perform additional tasks such as object detection \cite{Huang.2016} and semantic segmentation \cite{Long.2015}. Several network meta-architectures for object detection exist, which build upon a variety of convolutional layer designs for feature extraction. In terms of real-time application, single shot architectures have been shown to perform accurately while keeping computational times reasonably low \cite{Liu.2016}. In recent years, new feature extraction architectures have been proposed which increase the object detection performance when employed in a given meta-architecture \cite{He.2016,Redmon.2016,Ren.2015,Simonyan.2015}. Recently, further studies emerged to automatically fine-tune an initial neural network design to increase the detection performance or minimize the run-time, without effecting the detection performance significantly \cite{He.2018, Tan.2019}.

The success of neural networks for image data processing has led to an adaption to additional sensor principles and to sensor fusion. By incorporating multi-modal sensor data in the sensor fusion, researchers aim to obtain more reliable results for the different tasks involved in environmental perception for autonomous vehicles. \cite{Ku.2018} projects lidar data onto the 2D ground plane as a bird's-eye view and fuse it with camera data to perform 3D object detection. \cite{Chen.2016} projects the lidar onto the ground plane and onto a perpendicular image plane, and fuses both representations with the camera image in a neural network for object detection. \cite{Caltagirone.2018} fuses lidar and camera data in a neural network to segment the driveable road. The paper proposes a network structure which consists of two branches for lidar and camera input. The interconnections of these branches are trainable so that the network can learn an optimized depth level in the network for the data fusion during the training process. \cite{Yu.2019} uses a similar fusion approach while operating with a bird's-eye view projection for both camera and lidar.

Convolutional neural networks are widely applied to operate on regular 2D grids (e.g. images) or 3D grids (e.g. voxels). The 3D lidar object detection approaches discussed above apply the idea of transforming unstructured lidar point clouds onto a regular grid before feeding it into a neural network. We employ the same process to the radar data.

\cite{Ji.2008} uses radar detections to create regions of interest in camera images, in order to classify objects in these regions using a simple neural network. A similar approach of using the radar to guide the object detection in the image space is performed in a series of other works \cite{Kocic.2018,Han.2016,Zhang.2019,Zeng.01.08.2012}. \cite{Jha.2019} fuse independent detections of the camera and radar in order to associate the distance measurements of the radar with objects in the image space. \cite{Kim.2017} fuses independently tracked detections of each sensor to generate one final position estimation which incorporates the readings of both sensors. \cite{Lekic.2019} present a deep learning approach with Generative Adversarial Networks (GANs) to fuse camera data and radar data, incorporated into a 2D bird's-eye view grid map, in order to perform free space detection.

\cite{DiFeng.2019} gives an overview of deep learning methods for sensor fusion. They conclude that raw level fusion methods for image and radar data have merely been investigated to date, and that more research needs to be conducted in this respect. 
\cite{Chadwick.2019} projects low level radar data onto a camera image plane perpendicular to the road, and proposes a neural network for the fusion with the camera image. They use the range and the range rate of the radar as additional image channels. The paper proposes two fusion strategies by concatenation or by element-wise addition on a fixed layer after initial separated layers for the sensors. They show the benefit of the fusion strategy for a self-recorded dataset.

In this paper, we use a similar projection approach to \cite{Chadwick.2019} to project the radar data onto the vertical plane of the camera image with which it is fused. We propose a fusion network that is able to learn the network depth at which the fusion is most beneficial to reduce the network loss. We operate in the image space to operate with 2D ground-truth data which significantly facilitates training data generation in comparison to 3D labels. 

Due to the range rate measurement, moving objects can be distinguished from their surroundings in the radar data. For practical applications such as Adaptive Cruise Control (ACC), filtering for moving objects is applied to reduce the amount of false postives in the radar returns. At the same time, important stationary objects, e.g. cars stopped in front of a traffic lights, will be filtered out as well. In this approach no filtering for moving objects is performed, so that we are able to detect stationary and moving traffic objects alike.


\section{Radar Data Preprocessing}
\label{sec:radar_processing}

This section describes the projection of the radar data to the image plane that is used in our fusion approach. We describe the spatial calibration of the camera and radar sensors, how to deal with the missing height information from the radar returns, how to deal with the sparsity of the radar data, and ground-truth filtering methods to reduce the noise or clutter in the radar data.

The radar sensor outputs a sparse 2D point cloud with associated radar characteristics. The data used for this work includes the azimuth angle, the distance and the radar cross section (RCS). We transform the radar data from the 2D ground plane to a perpendicular image plane. The characteristics of the radar return are stored as pixel values in the augmented image. At the location of image pixels where no radar returns are present, the projected radar channel values are set to the value 0. The input camera image consists of three channels (red, green, blue); to this we add the aforementioned radar channels as the input for the neural network. In our own dataset, the field of view (FOV) of three radars overlap with the FOV of the front-facing fish-eye camera. We concatenate the point clouds of the three sensors into one and use this as the projected radar input source. The projection differs, as the nuScenes dataset uses a $\SI{70}{\degree}$ FOV camera while the TUM dataset uses a $\SI{185}{\degree}$ FOV fish-eye camera. In the nuScenes dataset, camera intrinsic and extrinsic mapping matrices are provided to transform a point from world coordinates into image coordinates. The non-linearities of a fish-eye lens cannot be mapped with a linear matrix operation. We use the calibration method presented by \cite{Scaramuzza.2006} to map the world coordinates to the image coordinates for our own data.

The radar detections give no information about the height at which they were received, which increases the difficulty to fuse the data types. The 3D coordinates of the radar detections are assumed to be returned from the ground plane that the vehicle is driving on. The projections are then extended in perpendicular direction to this plane, so as to account for the vertical extension of the objects to be detected. We detect traffic objects which can be classified as cars, trucks, motorcycles, bicycles and pedestrians. To cover the height of such object types, we assume a height extension of the radar detections of $\SI{3}{\metre}$ to associate camera pixels with radar data. The radar data is mapped with a pixel width of one into the image plane.

\begin{figure}[htbp]
    
    \begin{subfigure}[c]{88mm}
        \includegraphics[width=88mm]{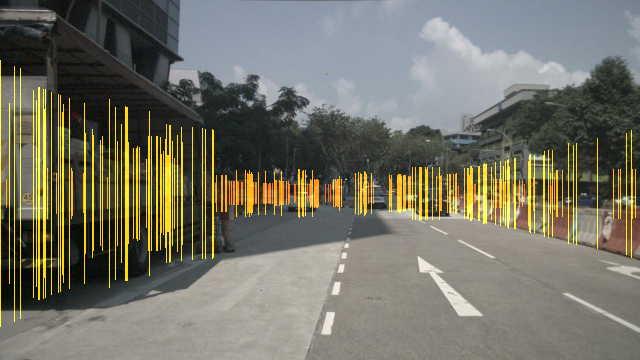}
        \subcaption{Without ground-truth noise filter}
        \label{fig:unfiltered_projection}
    \end{subfigure}

    \begin{subfigure}[c]{88mm}
        \includegraphics[width=88mm]{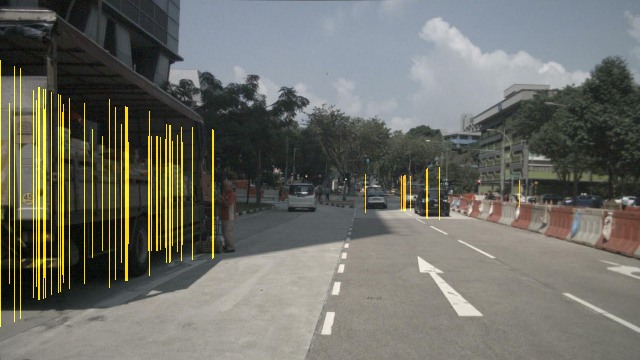}
        \subcaption{With ground-truth noise filter}
        \label{fig:filtered_projection}
    \end{subfigure}

	\caption{nuScenes sample with radar projection to the image plane for the last 13 radar cycles. Radar channels are shown in yellow. The red color shift depicts increasing distances. Best viewed in color.}
	\label{fig:radar_projection}
\end{figure}

The camera data in the nuScenes dataset is captured at a resolution of $\num{1600 x 900} = 1,440,000$ pixels at an opening angle of $\SI{70}{\degree}$ for the front camera. The lidar returns up to $14,000$ points for the same horizontal opening angle \cite{Geiger.2013}. On a fraction of the nuScenes dataset (nuScenes mini), we calculated an average of 57 radar detections per cycle for the front radar. The greater variety in the density of the radar and the camera data - in comparison to the lidar and the camera - poses the challenge of finding a suitable way to fuse the data in one shared network structure. For our own dataset, we use the  Continental ARS430 radar which has a different output format but comparable radar characteristics to the radar used in nuScenes. To deal with the sparsity of radar data, \cite{Lekic.2019} uses probabilistic grid maps to generate continuous information from the radar. In this work, we increase the density of radar data by jointly fusing the last 13 radar cycles (around $ \SI{1}{\second}$) to our data format. Ego-motion is compensated for this projection method. Target-vehicle motion cannot be compensated. The fusion of previous time steps adds to the information density of the radar input. At the same time, it can also add noise to the input data as the detections of moving objects at previous time steps do not align with the current object position. This drawback is tolerated to obtain an information gain due to the additional data. Figure \ref{fig:unfiltered_projection} shows the input data format for the neural network in an exemplary scene. The radar channels (distance and RCS) are mapped to the same locations and therefore shown in a uniform color.

The radar returns many detections coming from objects which are not relevant for the driving task, such as ghost objects, irrelevant objects and ground detections. These detections are called clutter or noise for the task at hand. In the evaluation, we compare the fusion of the raw noisy radar data with two additionally filtered approaches. First, in the nuScenes dataset, only a fraction of the labeled objects is detected by the radar. In training and evaluation, we therefore apply an annotation filter (AF), so that the filtered ground-truth data only contains objects which yield at least one radar detection. This is done via associating the 3D bounding boxes with radar points. The fusion approach is expected to show its potential for those objects which are detectable in both modalities. Second, we apply a ground-truth filter to the radar data which removes all radar detections outside of the 3D ground-truth bounding boxes. Of course, this step cannot be performed if applied to a real scenario. It is employed here to show the general feasibility of the fusion concept with less clutter in the input signal. The radar data after the application of the filter is shown in Figure~\ref{fig:filtered_projection}. Note, that the ground-truth radar filter (GRF) does not output perfect radar data and partly filters out relevant detections from the data for four reasons. First, we do not compensate the motion of other objects when we concatenate the past radar detections in the input. As the nuScenes dataset is labeled at \SI{2}{\hertz}, no ground-truth is available for intermediate radar detection cycles, radar object detections only present in intermediate cycles are possibly filtered out. Second, slight spatial miscalibrations between the radar and camera sensors result in a misalignment of the radar detection locations and the ground-truth bounding boxes at greater distances. Third, the data from the radar and the camera are not recorded at the exact same time. This leads to a spatial misalignment for moving objects. As we jointly operate on the last 13 detections of the radar, this effect is increased. Fourth, while the radar distance measurement is very reliable, its measurements are not perfect and slight inaccuracies can cause the detections to lie outside of the ground-truth bouding boxes. The unintended filtering of relevant data can partly be seen in Figure \ref{fig:filtered_projection}. In Section \ref{sec:evaluation}, we compare the results for the network using raw radar data and ground-truth filtered radar data. For the training and evaluation step, the 3D ground-truth bounding boxes are projected onto the 2D image plane.


\section{Network Fusion Architecture}
\label{sec:architecture}
Our neural network architecture builds on RetinaNet \cite{Lin.2017} as implemented in \cite{Fizyr.2015} with a VGG backbone \cite{Simonyan.2015}. The network is extended to deal with the additional radar channels of the augmented image. The output of the network is a 2D regression of bounding box coordinates and a classification score for the bounding box. The network is trained using focal loss, as proposed in \cite{Lin.2017}. Our baseline method uses a VGG feature extractor during the first convolutional layers.

\begin{figure}[htbp]
	\centering
\includegraphics[width=84mm]{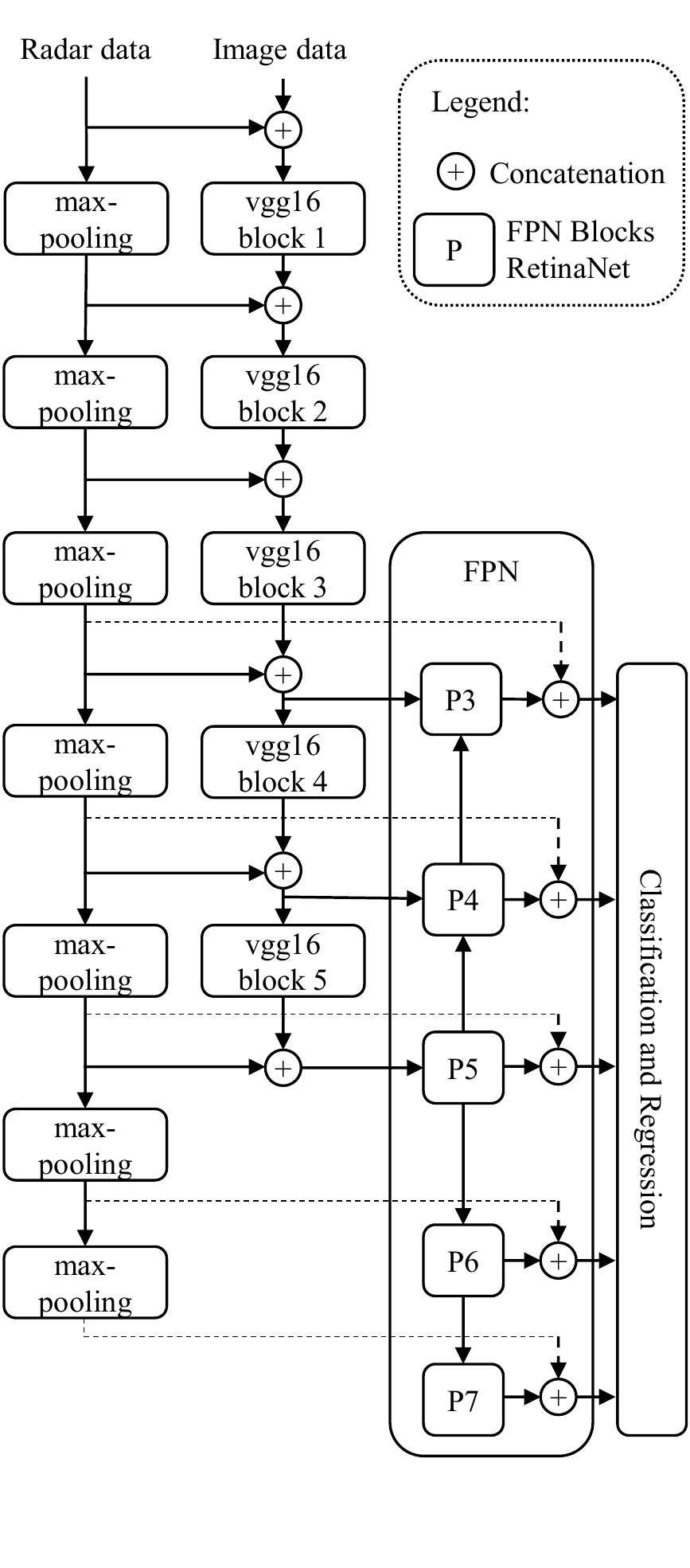}
	\caption{High-level structure of CameraRadarFusionNet (CRF-Net)}
	\label{fig:network_structure}
\end{figure}

The amount of information of one radar return is different from the information of a single pixel. The distance of an object to the ego-vehicle, as measured by the radar, can be considered more relevant to the driving task than a simple color value of a pixel of a camera. If both sensors are fused by concatenation in an early fusion, we should assume that the different data are semantically similar \cite{Liu.2018}. As we cannot strongly motivate this assumption, the fusion of the first layer of the network might not be optimal. In deeper layers of the neural network, the input data is compressed into a denser representation which ideally contains all the relevant input information. As it is hard to quantify the abstraction level of the information provided by each of the two sensor types, we design the network in a way that it learns itself at which depth level the fusion of the data is most beneficial to the overall loss minimization. The high-level structure of the network is shown in Figure \ref{fig:network_structure}. The main pipeline of the fusion network is shown in the center branch of the graph, composed of the VGG blocks. The camera and radar data is concatenated and fed into the network in the top row. This branch of the network is processed via the VGG layers for both the camera and radar data. In the left branch, the raw radar data is additionally fed into the network at deeper layers of the network at accordingly scaled input sizes through max-pooling. The radar data is concatenated to the output of the previous fused network layers of the main branch of the network. The Feature Pyramid Network (FPN) as introduced in \cite{Lin.2016} is represented by the blocks P3 through P7; herein, the radar channels are additionally fused by concatenation at each level. The outputs of the FPN blocks are finally processed by the bounding box regression and the classification blocks \cite{Lin.2017}. The optimizer implicitly teaches the network at which depth levels the radar data is fused with the greatest impact, by adapting the weights to the radar features at the different layers. A similar technique has been applied by \cite{Caltagirone.2018}.

We introduce a new training strategy to multi-modal sensor fusion for camera and radar data. The strategy is inspired by the technique Dropout \cite{Srivastava.2014}. Instead of single neurons, we simultaneously deactivate \textit{all} input neurons for the camera image data, for random training steps. This is done at a rate of $0.2$ of all training images. We call this technique \textit{BlackIn}. \cite{Ji.2016} introduced BlackOut which is inspired by dropout on the final layer of the network. The absence of camera input data pushes the network to rely more on the radar data. The goal is to teach the network the information value of the sparse radar data independently of the much denser camera representation. We begin the training with weights that are pretrained on images for the feature extractor. The training focus towards the radar, additionally intends to overcome this bias.

\section{Experiments and Results}
\label{sec:results}
In this section, we evaluate the network on the nuScenes dataset and a second dataset collected during the work for this paper. We compare our CameraRadarFusionNet (CRF-Net) with the baseline network, which is our adapted implementation of RetinaNet \cite{Lin.2017}. 

\subsection{Datasets}
\label{sec:datasets}
\paragraph{nuScenes dataset}
The nuScenes dataset is extensively described in \cite{Caesar.2019}. It is recorded in various locations and conditions in Boston and in Singapore. We condense the original 23 object classes into the classes shown in Table \ref{table:data_statistics} for our detection evaluation. The nuScenes results are evaluated with and without the application of ground-truth filters.

\begin{table}[htbp]
\caption{Number of objects per object class in nuScenes and our dataset}
\label{table:data_statistics}
\begin{center}
\begin{tabular}{|c|c|c|}
\hline
Object classes & nuScenes & TUM\\
\hline
Car & 22591 & 4020\\
Bus& 1332 & 109\\
Motorcycle& 729 & 10\\
Truck& 5015 & 14\\
Trailer& 1783 & 45\\
Bicycle& 616 & 438\\
Human& 10026 & 678\\
\hline
\end{tabular}
\end{center}
\end{table}

\paragraph{Our data (TUM)}
We utilize the same classes for evaluation as for the nuScenes dataset. Our dataset is annotated with 2D bounding boxes using the Computer Vision Annotation Tool (CVAT) \cite{Intel.2018}. As we lack 3D ground-truth data, no additional ground-truth filter can be applied to this dataset during the training and validation step. We reduce the default anchor sizes of RetinaNet by a factor two for our dataset, as the objects appear smaller on the fish-eye images.

\subsection{Training}
\label{sec:training}

We create an 60:20:20 split from the raw data of nuScenes to balance the amount of day, rain and night scenes in the training, validation and test set. We use the nuScenes images at an input size of 360~x~640 pixels. The fish-eye images of our dataset are processed at 720~x~1280 pixel resolution. Objects generally appear smaller in the fish-eye images which we want to compensate with the augmentation of the resolution. We weight the object classes according to the number of appearances in the respective datasets for the mean Average Precision (mAP) calculation. 

The weights of the VGG feature extractor are pretrained on the Imagenet dataset \cite{Deng.2009}. During preprocessing, the camera image channels are min-max scaled to the interval [-127.5,127.5], the radar channels remain unscaled. We perform data augmentation on our dataset because the amount of labeled data is relatively small.
The number of objects per class for each dataset is shown in Table \ref{table:data_statistics}.

Training and evaluation are performed with an Intel Xeon Silver 4112 CPU, 96GB RAM and a NVIDIA Titan XP GPU. 
On the nuScenes dataset, the networks are trained for 25 epochs and a batch size of 1 in a period of about 22 hours for the baseline network and about 24 hours for the CRF-Net. On our dataset, the networks are trained for 50 epochs and a batch size of 1 over a period of about 18 hours.   


\subsection{Evaluation}
\label{sec:evaluation}
Table \ref{table:mAP} shows the mean average precision for different configurations of our proposed network. The first block shows the results on the nuScenes dataset. The fusion network is achieving comparable but slightly higher detection results than the image network for the raw data inputs. The CRF-Net trained with BlackIn achieves a mAP of 0.35\,\%-points more than without BlackIn. In the next step, we apply the annotation filter (AF) which considers only objects which are detected by at least one radar point. When the network additionally learns on ground-truth filtered radar data (AF, GRF), the mAP advantage of the CRF-Net rises to 12.96\,\%-points compared to the image baseline (AF). The last line of the nuScenes block shows an additional comparison study. The radar channels are reduced to one channel which indicates solely the existence or non-existence of a radar detection in the image plane. The drop in the mAP score shows that the radar meta data, e.g. distance and RCS, are important for the detection result.

The second block of Table \ref{table:mAP} shows the data for our own dataset. The performance gain of the fusion network compared to the baseline (1.4\,\%-points) is greater for our data than for nuScenes. This could be due to the use of three partly overlapping radars in our data and due to the use of a more advanced radar sensor. In addition, we labeled objects that appear small in the images in our dataset; in the nuScenes dataset, objects at a distance greater than  \SI{80}{\meter} are mostly not labeled. As suggested in \cite{Chadwick.2019}, it is possible that the radar is beneficial especially for objects at a greater distance from the ego vehicle. The camera data differs in both datasets due to the different lens characteristics and the different input resolutions, so that a definite reason cannot be given here.

\begin{figure}[htbp]
    
    \begin{subfigure}[c]{85mm}
        \includegraphics[width=85mm]{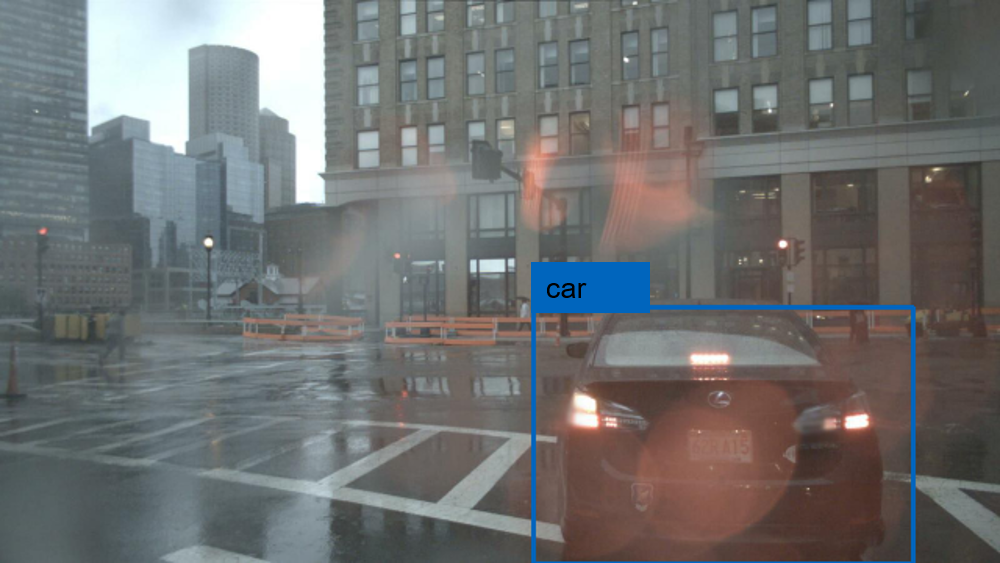}
        \subcaption{Baseline network detection}
    \end{subfigure}

    \begin{subfigure}[c]{85mm}
        \includegraphics[width=85mm]{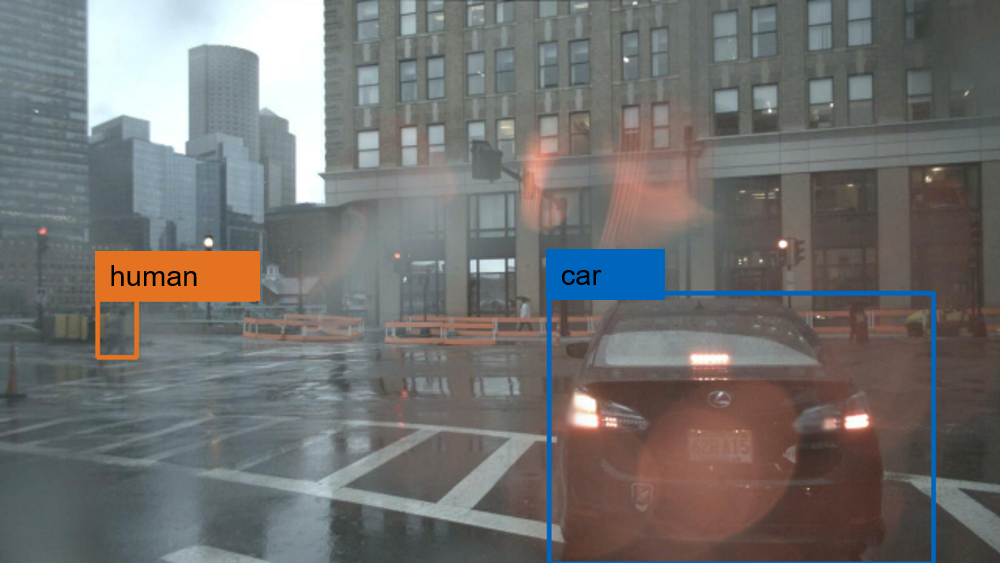}
        \subcaption{CRF-Net detection}
    \end{subfigure}

	\caption{Detection comparison of the baseline network (a) and the CRF-Net (b). The baseline network does not detect the pedestrian on the left.}
	\label{fig:radar_beneficial}
\end{figure}

Figure \ref{fig:radar_beneficial} qualitatively illustrates the superiority of the object detection with the CRF-Net for an example scene.


\begin{table}[ht]
    \caption{mAP scores of the baseline network and our CameraRadarFusionNet. Configurations: (AF) - Annotation filter, (GRF) - ground-truth radar filter, (NRM) - No radar meta data}
    \label{table:mAP}
    \centering
    \setlength{\extrarowheight}{0.4ex}
    \begin{tabular}{|c|c|c|}
    \hline
    Data & Network & mAP \\ \hline\hline
     \multirow{3}{*}[-5.0ex]{\rotatebox[origin=c]{90}{nuScenes}}
     & Baseline image network & 43.47\,\%  \\
     & CRF-Net w/o BlackIn & 43.6\,\% \\
     & \textbf{CRF-Net} & \textbf{43.95}\,\% \\     \cline{2-3}
     & Baseline image network (AF) & 43.03\,\%  \\
     & CRF-Net (AF) & 44.85\,\%  \\ 
     & \textbf{CRF-Net (AF, GRF)} & \textbf{55.99\,\%} \\
     & CRF-Net (AF, GRF, NRM) & 53.23\,\% \\     \hline\hline
     \multirow{3}{*}[1.60ex]{\rotatebox[origin=c]{90}{TUM}}
     & Baseline image network & 56.12\,\%  \\
     & \textbf{CRF-Net} & \textbf{57.50}\,\%   \\ \hline
  \end{tabular}
\end{table} 

The overall higher mAPs for the fusion network compared to the baseline presented in Table \ref{table:mAP} show the potential of the fusion approach. This potential motivates further research towards an ideal network architecture for this type of fusion. The performance gain for ground-truth filtered radar data motivates the development of a non-ground-truth based filtering method for the radar data during preprocessing, or inside the neural network. In future work, we will continue research into filtering out noisy radar detections before feeding them into the fusion network, to improve the results for the application under real-world conditions.

The baseline network needs \SI{33}{\milli\second} for the processing of one image at a size of 360 x 640 pixels. The CRF-Net needs \SI{43}{\milli\second} for the processing of the corresponding fused data. Additionally the data processing for the radar projection and channel generation amounts to \SI{56}{\milli\second} of CPU time. The time needed for the processing of the ground-truth filters is negligible. In our TUM dataset, we input the data at a higher resolution, which results in increased execution times. The baseline network processing takes \SI{92}{\milli\second}, the CRF-Net needs \SI{103}{\milli\second}, the data generation takes \SI{333}{\milli\second}. In this dataset more radar data is used and the projection is done with a fish-eye projection method which adds to the data generation time. However, the data generation is not optimized and the values are given as a reference to present the current status of the implementation.


\section{Conclusions and Outlook}
\label{sec:conlusions}
This paper proposes the CameraRadarFusion-Net (CRF-Net) architecture to fuse camera and radar sensor data of road vehicles. The research adapts ideas from lidar and camera data processing and shows a new direction for fusion with radar data. Difficulties and solutions to process the radar data are discussed. The BlackIn training strategy is introduced for the fusion of radar and camera data. We show that the fusion of radar and camera data in a neural network can augment the detection score of a state-of-the-art object detection network. This paper lends justification to a variety of areas for further research. As neural fusion for radar and camera data has only recently been studied in literature, finding optimized network architectures needs to be explored further. 

In the future, we plan research to design network layers to process the radar data prior to the fusion, so as to filter out noise in the radar data. The fusion with additional sensor modalities such as lidar data could further increase the detection accuracy, while at the same time adding complexity by augmenting the layers or through the need to introduce new design concepts. The study of the robustness of neural fusion approaches against spatial and temporal miscalibration of the sensors needs to be evaluated. We see an increased potential for multi-modal neural fusion for driving in adverse weather conditions. Additional datasets modeling these conditions need to be created to study this assumption. Lastly, as the radar sensor introduces distance information into the detection scheme, the applicability of the fusion concept to 3D object detection is a direction we want to explore. 

On the hardware side, high-resolution or imaging radars \cite{Brisken.2018} are expected to increase the information density of radar data and reduce the amount of clutter. The hardware advancement is expected to enable an increase in the detection results of our approach.


\section*{Contributions and Acknowledgments}
\label{sec:contributions}
Felix Nobis initiated the idea of this paper and contributed essentially to its conception and content. Maximilian Geisslinger and Markus Weber wrote their master theses in the research project and contributed to the conception, implementation and experimental results of this research. Johannes Betz revised the paper critically. Markus Lienkamp made an essential contribution to the conception of the research project. He revised the paper critically for important intellectual content. He
gave final approval of the version to be published and agrees to all aspects of the work.
As a guarantor, he accepts the responsibility for the overall integrity of the paper.
We express gratitude to Continental Engineering Service for funding for the underlying research project and for providing the sensor hardware and guidance for this research.

\bibliographystyle{./bibliography/IEEEtran} 
\bibliography{./bibliography/bibliography}

\end{document}